\documentclass[nohyperref]{article}

\usepackage{microtype}
\usepackage{graphicx}
\usepackage{subfigure}
\usepackage{booktabs}
\usepackage{multirow}

\usepackage{svg}

\usepackage{hyperref}

\usepackage[sectionbib]{natbib}
\usepackage{bibunits}

\usepackage[accepted]{icml2023}

\usepackage{amsmath}
\usepackage{amssymb}
\usepackage{mathtools}
\usepackage{amsthm}

\usepackage[capitalize,noabbrev]{cleveref}

\theoremstyle{plain}

\theoremstyle{definition}

\theoremstyle{remark}

\usepackage{tikz}
\usetikzlibrary{calc}
\usetikzlibrary{positioning}
\usetikzlibrary{decorations.markings}
\usetikzlibrary{fit}
\usetikzlibrary{shapes}

\definecolor{darkgreen}{rgb}{0.0, 0.5, 0.0}

\definecolor{UniBlue}{RGB}{0, 74, 153}
\definecolor{UniRed}{RGB}{193, 0, 42}
\definecolor{UniGrey}{RGB}{154, 155, 156}

\usepackage{enumitem}

\icmltitlerunning{Towards Automated Design of Riboswitches}

\defaultbibliographystyle{icml2023}
\defaultbibliography{bib/rna,bib/rna_data,bib/rna_folding,bib/dl,bib/rna_design,bib/automl}

\begin{document}

\twocolumn[
\icmltitle{Towards Automated Design of Riboswitches}
\icmlsetsymbol{equal}{*}

\begin{icmlauthorlist}
\icmlauthor{Frederic Runge}{uni}
\icmlauthor{Jörg K.H. Franke}{uni}
\icmlauthor{Frank Hutter}{uni}
\end{icmlauthorlist}

\icmlaffiliation{uni}{Department of Computer Science, University of Freiburg, Freiburg, Germany}

\icmlcorrespondingauthor{Frederic Runge}{runget@cs.uni-freiburg.de}
\icmlkeywords{Machine Learning, ICML}

\vskip 0.3in
]

\printAffiliationsAndNotice{}

\begin{abstract}
Experimental screening and selection pipelines for the discovery of novel riboswitches are expensive, time-consuming, and inefficient.
Using computational methods to reduce the number of candidates for the screen could drastically decrease these costs.
However, existing computational approaches do not fully satisfy all requirements for the design of such initial screening libraries.
In this work, we present a new method, \emph{libLEARNA}, capable of providing RNA focus libraries of diverse variable-length qualified candidates. Our novel structure-based design approach considers global properties as well as desired sequence and structure features.
We demonstrate the benefits of our method by designing theophylline riboswitch libraries, following a previously published protocol, and yielding 30\% more unique high-quality candidates.
\end{abstract}

\section{Introduction}
\label{sec:intro}
Riboswitches~\citep{mironov2002sensing} are regulatory RNA elements, typically located in the 5$\text{}^\prime$ untranslated region of messenger RNAs (mRNAs), that can specifically bind certain metabolites (ligand) to alter gene expression.
They consist of an \emph{aptamer} for recognizing the ligand, closely connected to an \emph{expression platform} that couples ligand binding with gene regulation.
The binding signal is transduced by a conformational change of the switch.
Riboswitches are typically found in bacteria but can also be artificially constructed.
Since aptamers are capable of binding nearly every molecular and supramolecular target~\citep{vorobyeva2018key}, riboswitches have become interesting tools for different applications including modulation of cellular functions~\citep{hallberg2017engineering} or the development of biosensors~\citep{findeiss2017design}.

However, the discovery of new functional riboswitches and aptamers is a challenging process.
Current experimental screening procedures, e.g. systematic evolution of ligands by exponential enrichment (SELEX)~\citep{tuerk1990systematic}, use RNA libraries of up to $10^{16}$ fixed-length random sequences~\citep{vorobyeva2018key} for a single target, resulting in expensive, time-consuming, and inefficient screens.
It is common knowledge that the length, the secondary structures, their diversity, and the nucleotide composition of the sequences in the initial library are crucial for a successful SELEX~\citep{vorobyeva2018key, kohlberger2022selex}.
However, random libraries of fixed-length sequences are still most widely used.
Focused design can help to decrease the size of the initial library and therefore drastically reduce the costs of these SELEX pipelines.

To tackle the problem of library generation, there exists different computational approaches.
While earlier work focused on candidate selection from fixed-length random libraries~\citep{kim2010computational, zhou2015searching} based on different features, more recent approaches seek to tackle the problem with generative algorithms, employing recurrent neural networks~\citep{im2019generative}, Restricted Boltzmann Machines~\citep{di2022generative}, variational autoencoders (VAE)~\citep{iwano2022generative}, Monte-Carlo Tree Search\citep{lee2021predicting}, or evolutionary algorithms~\citep{andress2022daptev}.
However, these methods require initial libraries either from previous SELEX data~\citep{di2022generative, iwano2022generative}, of known good binders~\citep{andress2022daptev}, or even known interaction scores~\citep{im2019generative, lee2021predicting}, which is often unavailable for the discovery of novel aptamers.
Further, the design process works with sequence information only, ignoring nucleotide compositions or using secondary structure only to enable docking simulations for scoring.

As an alternative, rational structure-based RNA design algorithms have previously been employed to successfully design riboswitches~\citep{wachsmuth_2012, ender2021ligand}.
The general goal of these algorithms is to find an RNA sequence that folds into a desired secondary structure.
However, all structure-based RNA design algorithms are either limited to the design of fixed-length candidates~\citep{hofacker_1994, andronescu_2004, taneda2010modena, kleinkauf_2015, eastman_2018, runge_2019} or require a fully defined pairing scheme of the input structure which is a strong restriction on the structural diversity and hinders their applicability to larger scale library designs.
This is also true for tools that were specifically developed for the task of riboswitch design, like the sampling approach proposed by \citet{hammer2017rnablueprint}.

In this work, we present a novel structure-based RNA design algorithm, \emph{libLEARNA}, that is capable of designing large amounts of diverse candidate sequences with different lengths, while considering desired sequence and structural constraints, as well as nucleotide distributions, during the design process.
In particular, our contributions are as follows:
\setlist{nolistsep}
\begin{itemize}[noitemsep,topsep=-0.5em]
  \item We introduce a novel \textit{RNA design paradigm} that enables the design of variable-length RNA sequences from desired arbitrary sequence and structure parts.
  \item We improve an existing \textit{RNA design algorithm}, \emph{LEARNA}~\citep{runge_2019}, with a masked training objective to enable efficient RNA library generation, including global sequence properties due to the robustness of our approach.
  \item We show the benefits of our approach by exemplarily designing theophylline riboswitch libraries, following a previously proposed protocol by \citet{wachsmuth_2012}, yielding 30\% more unique candidates compared to the original approach and up to 47\% when designing sequences with desired G and C nucleotide ratios, while providing greater structural diversity across nearly uniformly distributed sequence lengths.
\end{itemize}

\section{Method}
\label{sec:method}
We first recap the foundation of our method, the \emph{LEARNA} algorithm, followed by the changes we apply to the state representations, the training procedure, and the meta-optimization process which leads to its successor \emph{libLEARNA}.

\subsection{Background}
\label{sec:background}
\emph{LEARNA} is a generative automated deep reinforcement learning (AutoRL)~\citep{parker-holder-jair22a} algorithm for the inverse RNA folding problem~\citep{hofacker_1994}. The algorithm samples RNA sequences from a learned policy given a target secondary structure.
In an inner-loop, a deep reinforcement learning (RL) algorithm meta-learns an RNA design policy across thousands of different inverse RNA folding tasks. The validation loss is communicated to an efficient Bayesian Optimization method, BOHB~\citep{falkner_2018}, which jointly optimizes the configuration of the RL system in the outer-loop.
The actions of the RL agent correspond to placing a nucleotide (A, C, G, or U) for each position, or directly placing Watson-Crick pairs (A-U, U-A, G-C, or C-G) in case the structure indicates that the given position is paired to another nucleotide.
States are defined as local representations of the input structure in dot-bracket format~\citep{hofacker_1994} using an n-gram centered around the current position.
After applying a folding algorithm to the designed sequence, the reward function is based on the Hamming distance between the folded candidate and the desired structure.
Finally, the best-performing configuration on the validation set is evaluated at test time.

\begin{table*}[t]
\centering
\caption[Riboswitch Re-design]{\textbf{Originally proposed theophylline riboswitch constructs and partial RNA design space formulation.} (Top) The sequence parts of the six originally proposed riboswitch constructs by \citet{wachsmuth_2012} and the sequence part of the design space. (Bottom) The corresponding structure parts of the constructs and the structural part of the design space. Red: TCT8-4 theophylline aptamer; green: variable length spacer domain; blue: domain ought to pair with the aptamer (complementary to the 3$^{\prime}$-end of the aptamer sequence in the original design); black: 8-U-stretch. Masked positions are indicated with \texttt{?}, positions for extensions are indicated with $\hat{?}$.}\vskip 0.15in
\begin{center}
\begin{small}
\begin{sc}
\resizebox{\linewidth}{!}{%
\begin{tabular}{lllll}
\toprule
\textbf{Construct} & \textbf{Aptamer} & \textbf{Spacer} & \textbf{Complementary Region} & \textbf{8-U-Stretch} \\ \midrule
RS1 Sequence       & \texttt{\color{UniRed}AAGUGAUACCAGCAUCGUCUUGAUGCCCUUGGCAGCACUUCA} & \texttt{\color{darkgreen}UUACAUC} & \texttt{\color{UniBlue}UGAAGUGCUGCC} & \texttt{\color{black}UUUUUUUU} \\
RS2 Sequence       & \texttt{\color{UniRed}AAGUGAUACCAGCAUCGUCUUGAUGCCCUUGGCAGCACUUCA} & \texttt{\color{darkgreen}UGAUCUCGCU} & \texttt{\color{UniBlue}UGAAGUGCUGC} & \texttt{\color{black}UUUUUUUU} \\
RS3 Sequence       & \texttt{\color{UniRed}AAGUGAUACCAGCAUCGUCUUGAUGCCCUUGGCAGCACUUCA} & \texttt{\color{darkgreen}UUUACAUACUCGGUAAAC} & \texttt{\color{UniBlue}UGAAGUGCUGCCA} & \texttt{\color{black}UUUUUUUU} \\
RS4 Sequence       & \texttt{\color{UniRed}AAGUGAUACCAGCAUCGUCUUGAUGCCCUUGGCAGCACUUCA} & \texttt{\color{darkgreen}AACCGAAAUUUGCGCU} & \texttt{\color{UniBlue}UGAAGUGCUGC} & \texttt{\color{black}UUUUUUUU} \\
RS8 Sequence       & \texttt{\color{UniRed}AAGUGAUACCAGCAUCGUCUUGAUGCCCUUGGCAGCACUUCA} & \texttt{\color{darkgreen}CUCCUAGUGGAG} &  \texttt{\color{UniBlue}UGAAGUGCUG} & \texttt{\color{black}UUUUUUUU} \\
RS10 Sequence      & \texttt{\color{UniRed}AAGUGAUACCAGCAUCGUCUUGAUGCCCUUGGCAGCACUUCA} & \texttt{\color{darkgreen}GAAAUCUC} & \texttt{\color{UniBlue}UGAAGUGCUG} & \texttt{\color{black}UUUUUUUU} \\
Task Sequence Parts & \texttt{\color{UniRed}AAGUGAUACCAGCAUCGUCUUGAUGCCCUUGGCAGCACUUCA} & \texttt{\color{darkgreen}$\hat{?}$??????$\hat{?}$} & \texttt{\color{UniBlue}UGAAGUGCUG$\hat{?}$} & \texttt{\color{black}UUUUUUUU} \\ \midrule
RS1 Structure      & \texttt{\color{UniRed}...........(((((.....)))))....((((((((((((} & \texttt{\color{darkgreen}.......} & \texttt{\color{UniBlue}))))))))))))} & \texttt{\color{black}........} \\
RS2 Structure      & \texttt{\color{UniRed}...........(((((.....)))))....((((((((((((} & \texttt{\color{darkgreen}..........} & \texttt{\color{UniBlue})))))))))))} & \texttt{\color{black}).......} \\
RS3 Structure      & \texttt{\color{UniRed}...........(((((.....)))))...(((((((((((((} & \texttt{\color{darkgreen}(((((.......))))).} & \texttt{\color{UniBlue})))))))))))))} & \texttt{........} \\
RS4 Structure      & \texttt{\color{UniRed}...........(((((.....)))))....((((((((((((} & \texttt{\color{darkgreen}(..((.......)).)} & \texttt{\color{UniBlue})))))))))))} & \texttt{).......} \\
RS8 Structure      & \texttt{\color{UniRed}........((((((((.....)))))...)))((((((((((} & \texttt{\color{darkgreen}((((....))))} & \texttt{\color{UniBlue}))))))))))} & \texttt{........} \\
RS10 Structure      & \texttt{\color{UniRed}........((((((((.....)))))...)))((((((((((} & \texttt{\color{darkgreen}((....)} & \texttt{\color{UniBlue})))))))))))} & \texttt{........} \\
Task Structure Parts & \texttt{\color{UniRed}........???(((((.....)))))...???((((((((((} & \texttt{\color{darkgreen}$\hat{?}$??....$\hat{?}$} & \texttt{\color{UniBlue}))))))))))$\hat{?}$} & \texttt{\color{black}?.......} \\
\bottomrule
\end{tabular}%%
}
\end{sc}
\end{small}
\end{center}
\vskip -0.1in
\label{tbl:merged_task}
\end{table*}

\subsection{libLEARNA}
\label{sec:libLEARNA}
We improve \emph{LEARNA} to design RNA candidates from arbitrary sequence and structure constraints due to an extended state representation, changes in the training procedure, and additional dimensions, as well as changes to the general objective of the meta-optimization process.

\textbf{State Representations}\hspace{1em} To inform \emph{libLEARNA} about constraints in the sequence and the structure, we extend the state representations of \emph{LEARNA}. In particular, we numerically encode pairs of a sequence symbol and its corresponding structure symbol for each position of the design task. This enlarges the state space of \emph{libLEARNA} compared to \emph{LEARNA} but allows \emph{libLEARNA} to be more flexible in terms of design tasks it can be applied to.

\label{par:training}
\textbf{Training}\hspace{1em} Instead of training on tasks of the inverse RNA folding problem, we use a masked training objective similar to the masked language model training in BERT~\citep{devlin2018bert}.
In particular, we follow \citet{runge_2019} to generate three training datasets with different length distributions ($\leq 200$ nucleotides (nt), $\geq 200$ nt, random length) of $100000$ samples each, and a non-overlapping validation set of $100$ samples from the Rfam~\citep{rfam_2003} database version 14.1 using \emph{RNAfold}~\citep{lorenz_2011} for folding the sequences.
However, in contrast to \citet{runge_2019} we do not mask the entire sequences but derive tasks that correspond to the design of RNAs from desired sequence and structure parts by applying a random masking procedure to the sequences and the structures detailed in Appendix~\ref{app:data}.
The task of \emph{libLEARNA} during training then is to fill the masked parts of the sequence such that, after applying a folding algorithm to the designed sequence, all positions of the resulting folding satisfy the given positional constraints of the masked structure.
Similar to \citet{runge_2019}, the reward is based on the Hamming distance, while masked positions in the structure are ignored.
Our design procedure describes a completely new structure-based RNA design paradigm since the design algorithm is not informed about the pairing conditions \emph{a priori}. This is in contrast to previous work in the field of structure-based RNA design and enables the design of RNAs from arbitrary sequence and structure parts while creating larger structural diversity and further allowing us to extend the task at any given point since we are no longer bound to explicit pairing positions.
However, we allow indexing pairs to explicitly indicate pairing positions if desired.
More details and examples of tasks for the datasets can be found in Appendix~\ref{app:data}.

\textbf{Meta-Optimization}\hspace{1em} For \emph{libLEARNA}, we mainly adopt the configuration space proposed by \citet{runge_2019} but introduce four new dimensions. (1) While algorithms for the inverse RNA folding problem typically benefit from directly predicting Watson-Crick base pairs at sites that are known to be paired with another nucleotide, the pairing partners for many paired sites are not trivially distinguishable as a result of our masking procedure during training. However, we include the choice to make use of the direct prediction of pairs if the pairing partner can be identified, i.e. there is no masking between the pairing positions. (2) We add a choice to the configuration space to dynamically adapt the states based on the design progress to inform the agent about its own decisions. (3) The choice of training data and (4) the schedule of the tasks during training can have a strong impact on the final performance of the learning algorithm. We, therefore, include two dimensions to choose from three different training data distributions and curricula (unsorted tasks or sorted by length).
We further allowed searching over an additional LSTM layer.
The result is an 18-dimensional search space to jointly optimize over the network architecture, all parts of the MDP, as well as training hyperparameters, task distributions, and schedule, using BOHB~\citep{falkner_2018}.
We use the exact same setup during meta-optimization as~\citet{runge_2019}, with the same training budgets and validation protocols.
However, while \citet{runge_2019} optimized for an RL algorithm without any policy updates at test time, we directly optimize for an algorithm with policy updates at evaluation time to increase the adaptation capabilities of our approach.
The meta-optimization procedure, the configuration space, as well as the final configuration of \emph{libLEARNA} are detailed in Appendix~\ref{app:meta_optimization}.

\section{Riboswitch Library Generation}
\label{sec:lib_generation}
After describing the procedure of the original design and evaluation of candidates for synthetic riboswitches for theophylline-dependent regulation of transcription proposed by \citet{wachsmuth_2012} we detail our approach for the construction of an RNA design space and the design of RNA libraries and evaluate both methods.

\subsection{Original Setup}
\label{sec:riboswitch_original_setup}

\paragraph{Initial Setting}
Originally, \citet{wachsmuth_2012} constructed riboswitch candidates from (1)~the TCT8-4 theophylline aptamer sequence and structure, (2)~a spacer sequence of 6 to 20 nucleotides (nt), (3)~a sequence of 10nt to 21nt complementary to the 3$^{\prime}$-end of the aptamer, and (4)~ a U-stretch of 8nt at the 3$^{\prime}$-end of the construct.

\paragraph{Design Procedure} To generate candidates, \citet{wachsmuth_2012} designed a large library of random sequences for the spacer region (6-20nt) and a library of sequences complementary to the 3$^{\prime}$-end of the aptamer (10-21nt).
From these sets, randomly sampled sequences were combined with the aptamer and the 8-U-stretch.

\subsection{libLEARNA Setup}

\paragraph{Design Space Formulation} We start the formulation of our design space from the entire sequence and the structure part of the unbound aptamer, a spacer region of unknown sequence, a region complementary to the 3$\text{}^\prime$-end of the aptamer of at least 10nt and the unpaired 8-U-stretch, similar to \citet{wachsmuth_2012}.
To ensure a length of 6nt of the spacer region, we use the shared structure constraints of the six final constructs proposed by \citet{wachsmuth_2012}, which were also used to restrict the unknown parts of the aptamer.
We then introduce three positions where the task can be extended (indicated by \texttt{$\hat{?}$} in Table~\ref{tbl:merged_task}) to fit the requirements and fix the length of the design space to 66nt-91nt according to \citet{wachsmuth_2012}.
The final definition of the design space and the proposed constructs of \citet{wachsmuth_2012} are shown in Table~\ref{tbl:merged_task}.

\paragraph{Design Procedure} The input to \emph{libLEARNA} is the design space shown in Table~\ref{tbl:merged_task}. To generate candidates, \emph{libLEARNA} internally first samples a masked task from the design space by uniformly sampling a sequence of masking tokens for each extension position considering the length constraints and then tries to solve the task with a single shot.

\begin{table}[t]
\centering
\caption[Riboswitch design]{\textbf{Overview of candidates that satisfy the design criteria.} All numbers are averages across five runs with different random seeds.}\vskip 0.15in
\begin{center}
\begin{small}
\begin{sc}
\resizebox{\linewidth}{!}{%
\begin{tabular}{lrr}
\toprule
\textbf{Method} & \textbf{Valid Candidates [\%]} & \textbf{Unique Structures} \\
\midrule
\citet{wachsmuth_2012} & 41,7 & 6316.6 \\
\emph{libLEARNA} & \textbf{70,9} & \textbf{8269.6} \\
\bottomrule
\end{tabular}%%
}
\end{sc}
\end{small}
\end{center}
\vskip -0.1in
\label{tbl:riboswitch_results}
\end{table}

\subsection{Experiments}
We assess the performance of \emph{libLEARNA} against the originally proposed library generation procedure proposed by \citet{wachsmuth_2012} in two experiments: (1) The design of a library based on sequence and structure constraints only, and (2) the design of candidates when additionally querying \emph{libLEARNA} to design candidates with a specific G and C nucleotide ratio (GC-content), given a tolerance of $0.01$.
We note that \emph{libLEARNA} was never trained for predictions with desired GC-contents but that a GC-content loss-term (the absolute deviation of the GC-content of the current candidate sequence from the desired GC-content) was simply added to the reward function of \emph{libLEARNA}. The reward function then is a weighted sum (without any tuning, setting all weights to 1) of the structure- and the GC-loss. However, similar to the structural improvement step described by \citet{runge_2019}, we implement a GC-improvement step to guide the agent for this challenging task, detailed in Appendix~\ref{app:GIS}.
For each experiment, we generate 50000 candidates with the approach of \citet{wachsmuth_2012} and \emph{libLEARNA} and evaluate them as follows.

\paragraph{Evaluation} The evaluation of the designed candidates was performed following \citet{wachsmuth_2012}: after dropping duplicates, the designed sequences were folded using \emph{RNAfold} and verified for (1) the existence of two hairpin structure elements, the aptamer hairpin for binding the ligand and the terminator hairpin formed between the 3$\text{}^\prime$-end of the aptamer, the spacer and the region complementary to the aptamer sequence, (2) no pairing within the last seven nucleotides of the 8-U-stretch, (3) no pairing between the spacer and the aptamer in a sequence of folding steps, simulating co-transcriptional folding with a fixed elongation speed of five. A candidate that did not pass all criteria was rejected.

\begin{figure}[t]
\vskip 0.2in
\begin{center}
        \resizebox{0.7\linewidth}{!}{%
        \centerline{%
        \includegraphics{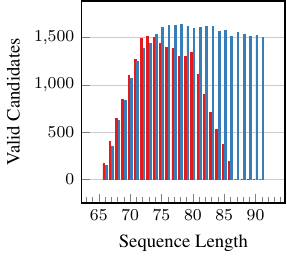}%
		\includegraphics{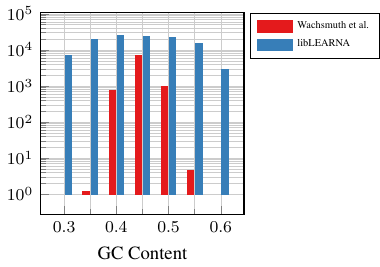}
        }}
\caption[GC-content and length performance, skip=0.1in]{
    \textbf{Length and GC-content distribution in the generated libraries.} (Left) The plot displays the distribution of the candidates across different lengths. (Right) The plot shows the number of unique candidates that satisfy all design criteria and have the specified G and C nucleotide distribution when running \emph{libLEARNA} with a desired GC-content. All numbers are averages across five runs with different random seeds.
	}	\label{fig:results}
\end{center}
\vskip -0.2in
\end{figure}

\paragraph{Results} We observe that \emph{libLEARNA} generates considerably more candidates that pass the design criteria compared to the original procedure proposed by \citet{wachsmuth_2012}, yielding 30\% more satisfying candidates on average (Table~\ref{tbl:riboswitch_results}). Further, the candidates are nearly uniformly distributed across the lengths of the design space, especially for the longer sequences (Figure~\ref{fig:results} left), and the structure diversity generated by \emph{libLEARNA} is around 23\% higher (Table~\ref{tbl:riboswitch_results}). When designing candidates with desired GC-contents, \emph{libLEARNA} provides up to 47\% more candidates that satisfy the design criteria and contain the specific G and C nucleotide distribution (Figure~\ref{fig:results} right). Remarkably, \emph{libLEARNA} can also design such candidates on the margins of possible GC-contents ($0.3$ and $0.6$) which lay between $0.29$ and $0.63$ for the given riboswitch design space.

\section{Conclusion}
\label{sec:conclusion}
We propose a new RNA design paradigm based on a masked prediction task that enables the design of RNA libraries with large structure diversity from arbitrary sequence and structure constraints. We use the paradigm and develop a new algorithm, \emph{libLEARNA}, capable of generating these libraries.
We exemplarily demonstrate the benefits of our method on the design of variable-length riboswitches, showing the efficiency of our approach.
Our results for library generation with desired G and C nucleotide distributions show that \emph{libLEARNA} can handle global constraints, while the robustness to reward changes suggests potential for handling more constraints, e.g. a desired energy threshold of the structures, which we will assess in future work.
Overall, our approach bears great potential to support experimental pipelines.

\section*{Acknowledgements} This research was funded by the Deutsche Forschungsgemeinschaft (DFG, German Research Foundation) under grant number 417962828. The authors further acknowledge support by the state of Baden-W\"{u}rttemberg through bwHPC and the German Research Foundation (DFG) through grant no INST 39/963-1 FUGG.

\bibliography{bib/rna,bib/rna_data,bib/rna_folding,bib/dl,bib/rna_design,bib/automl}

\begin{thebibliography}{27}
\providecommand{\natexlab}[1]{#1}
\providecommand{\url}[1]{\texttt{#1}}
\expandafter\ifx\csname urlstyle\endcsname\relax
  \providecommand{\doi}[1]{doi: #1}\else
  \providecommand{\doi}{doi: \begingroup \urlstyle{rm}\Url}\fi

\bibitem[Andress et~al.(2022)Andress, Kappel, Cuperlovic-Culf, Yan, and
  Li]{andress2022daptev}
Andress, C., Kappel, K., Cuperlovic-Culf, M., Yan, H., and Li, Y.
\newblock Daptev: Deep aptamer evolutionary modelling for covid-19 drug design.
\newblock \emph{bioRxiv}, pp.\  2022--11, 2022.

\bibitem[Andronescu et~al.(2004)Andronescu, Fejes, Hutter, Hoos, and
  Condon]{andronescu_2004}
Andronescu, M., Fejes, A.~P., Hutter, F., Hoos, H.~H., and Condon, A.
\newblock A new algorithm for rna secondary structure design.
\newblock \emph{Journal of molecular biology}, 336\penalty0 (3):\penalty0
  607--624, 2004.

\bibitem[Devlin et~al.(2018)Devlin, Chang, Lee, and Toutanova]{devlin2018bert}
Devlin, J., Chang, M.-W., Lee, K., and Toutanova, K.
\newblock Bert: Pre-training of deep bidirectional transformers for language
  understanding.
\newblock \emph{arXiv preprint arXiv:1810.04805}, 2018.

\bibitem[Di~Gioacchino et~al.(2022)Di~Gioacchino, Procyk, Molari, Schreck,
  Zhou, Liu, Monasson, Cocco, and {\v{S}}ulc]{di2022generative}
Di~Gioacchino, A., Procyk, J., Molari, M., Schreck, J.~S., Zhou, Y., Liu, Y.,
  Monasson, R., Cocco, S., and {\v{S}}ulc, P.
\newblock Generative and interpretable machine learning for aptamer design and
  analysis of in vitro sequence selection.
\newblock \emph{PLoS computational biology}, 18\penalty0 (9):\penalty0
  e1010561, 2022.

\bibitem[Eastman et~al.(2018)Eastman, Shi, Ramsundar, and Pande]{eastman_2018}
Eastman, P., Shi, J., Ramsundar, B., and Pande, V.~S.
\newblock Solving the rna design problem with reinforcement learning.
\newblock \emph{PLoS computational biology}, 14\penalty0 (6):\penalty0
  e1006176, 2018.

\bibitem[Ender et~al.(2021)Ender, Etzel, Hammer, Findei{\ss}, Stadler, and
  M{\"o}rl]{ender2021ligand}
Ender, A., Etzel, M., Hammer, S., Findei{\ss}, S., Stadler, P., and M{\"o}rl,
  M.
\newblock Ligand-dependent trna processing by a rationally designed rnase p
  riboswitch.
\newblock \emph{Nucleic acids research}, 49\penalty0 (3):\penalty0 1784--1800,
  2021.

\bibitem[Falkner et~al.(2018)Falkner, Klein, and Hutter]{falkner_2018}
Falkner, S., Klein, A., and Hutter, F.
\newblock {BOHB}: Robust and efficient hyperparameter optimization at scale.
\newblock In Dy, J. and Krause, A. (eds.), \emph{Proceedings of the 35th
  International Conference on Machine Learning}, volume~80 of \emph{Proceedings
  of Machine Learning Research}, pp.\  1437--1446, Stockholmsmässan, Stockholm
  Sweden, 10--15 Jul 2018. PMLR.

\bibitem[Findei{\ss} et~al.(2017)Findei{\ss}, Etzel, Will, M{\"o}rl, and
  Stadler]{findeiss2017design}
Findei{\ss}, S., Etzel, M., Will, S., M{\"o}rl, M., and Stadler, P.~F.
\newblock Design of artificial riboswitches as biosensors.
\newblock \emph{Sensors}, 17\penalty0 (9):\penalty0 1990, 2017.

\bibitem[Griffiths-Jones et~al.(2003)Griffiths-Jones, Bateman, Marshall,
  Khanna, and Eddy]{rfam_2003}
Griffiths-Jones, S., Bateman, A., Marshall, M., Khanna, A., and Eddy, S.~R.
\newblock {Rfam: an RNA family database}.
\newblock \emph{Nucleic Acids Research}, 31\penalty0 (1):\penalty0 439--441, 01
  2003.
\newblock ISSN 0305-1048.

\bibitem[Hallberg et~al.(2017)Hallberg, Su, Kitto, and
  Hammond]{hallberg2017engineering}
Hallberg, Z.~F., Su, Y., Kitto, R.~Z., and Hammond, M.~C.
\newblock Engineering and in vivo applications of riboswitches.
\newblock \emph{Annual review of biochemistry}, 86:\penalty0 515--539, 2017.

\bibitem[Hammer et~al.(2017)Hammer, Tschiatschek, Flamm, Hofacker, and
  Findei{\ss}]{hammer2017rnablueprint}
Hammer, S., Tschiatschek, B., Flamm, C., Hofacker, I.~L., and Findei{\ss}, S.
\newblock Rnablueprint: flexible multiple target nucleic acid sequence design.
\newblock \emph{Bioinformatics}, 33\penalty0 (18):\penalty0 2850--2858, 2017.

\bibitem[Hofacker et~al.(1994)Hofacker, Fontana, Stadler, Bonhoeffer, Tacker,
  and Schuster]{hofacker_1994}
Hofacker, I., Fontana, W., Stadler, P., Bonhoeffer, S., Tacker, M., and
  Schuster, P.
\newblock {F}ast {F}olding and {C}omparison of {RNA} {S}econdary {S}tructures.
\newblock \emph{Monatshefte fuer Chemie/Chemical Monthly}, 125:\penalty0
  167--188, 02 1994.

\bibitem[Im et~al.(2019)Im, Park, and Han]{im2019generative}
Im, J., Park, B., and Han, K.
\newblock A generative model for constructing nucleic acid sequences binding to
  a protein.
\newblock \emph{BMC genomics}, 20\penalty0 (13):\penalty0 1--13, 2019.

\bibitem[Iwano et~al.(2022)Iwano, Adachi, Aoki, Nakamura, and
  Hamada]{iwano2022generative}
Iwano, N., Adachi, T., Aoki, K., Nakamura, Y., and Hamada, M.
\newblock Generative aptamer discovery using raptgen.
\newblock \emph{Nature Computational Science}, 2\penalty0 (6):\penalty0
  378--386, 2022.

\bibitem[Kim et~al.(2010)Kim, Izzo, Elmetwaly, Gan, and
  Schlick]{kim2010computational}
Kim, N., Izzo, J.~A., Elmetwaly, S., Gan, H.~H., and Schlick, T.
\newblock Computational generation and screening of rna motifs in large
  nucleotide sequence pools.
\newblock \emph{Nucleic acids research}, 38\penalty0 (13):\penalty0 e139--e139,
  2010.

\bibitem[Kleinkauf et~al.(2015)Kleinkauf, Houwaart, Backofen, and
  Mann]{kleinkauf_2015}
Kleinkauf, R., Houwaart, T., Backofen, R., and Mann, M.
\newblock anta{RNA}--{M}ulti-objective inverse folding of pseudoknot {RNA}
  using ant-colony optimization.
\newblock \emph{BMC bioinformatics}, 16\penalty0 (1):\penalty0 389, 2015.

\bibitem[Kohlberger \& Gadermaier(2022)Kohlberger and
  Gadermaier]{kohlberger2022selex}
Kohlberger, M. and Gadermaier, G.
\newblock Selex: Critical factors and optimization strategies for successful
  aptamer selection.
\newblock \emph{Biotechnology and Applied Biochemistry}, 69\penalty0
  (5):\penalty0 1771--1792, 2022.

\bibitem[Lee et~al.(2021)Lee, Jang, Kang, and Song]{lee2021predicting}
Lee, G., Jang, G.~H., Kang, H.~Y., and Song, G.
\newblock Predicting aptamer sequences that interact with target proteins using
  an aptamer-protein interaction classifier and a monte carlo tree search
  approach.
\newblock \emph{PloS one}, 16\penalty0 (6):\penalty0 e0253760, 2021.

\bibitem[Lorenz et~al.(2011)Lorenz, Bernhart, H{\"o}ner~zu Siederdissen, Tafer,
  Flamm, Stadler, and Hofacker]{lorenz_2011}
Lorenz, R., Bernhart, S.~H., H{\"o}ner~zu Siederdissen, C., Tafer, H., Flamm,
  C., Stadler, P.~F., and Hofacker, I.~L.
\newblock Viennarna package 2.0.
\newblock \emph{Algorithms for Molecular Biology}, 6\penalty0 (1):\penalty0 26,
  Nov 2011.
\newblock ISSN 1748-7188.

\bibitem[Mironov et~al.(2002)Mironov, Gusarov, Rafikov, Lopez, Shatalin,
  Kreneva, Perumov, and Nudler]{mironov2002sensing}
Mironov, A.~S., Gusarov, I., Rafikov, R., Lopez, L.~E., Shatalin, K., Kreneva,
  R.~A., Perumov, D.~A., and Nudler, E.
\newblock Sensing small molecules by nascent rna: a mechanism to control
  transcription in bacteria.
\newblock \emph{cell}, 111\penalty0 (5):\penalty0 747--756, 2002.

\bibitem[Parker-Holder et~al.(2022)Parker-Holder, Rajan, Song, Biedenkapp,
  Miao, Eimer, Zhang, Nguyen, Calandra, Faust, Hutter, and
  Lindauer]{parker-holder-jair22a}
Parker-Holder, J., Rajan, R., Song, X., Biedenkapp, A., Miao, Y., Eimer, T.,
  Zhang, B., Nguyen, V., Calandra, R., Faust, A., Hutter, F., and Lindauer, M.
\newblock Automated reinforcement learning (autorl): A survey and open
  problems.
\newblock \emph{Journal of Artificial Intelligence Research (JAIR)},
  74:\penalty0 517--568, 2022.

\bibitem[Runge et~al.(2019)Runge, Stoll, Falkner, and Hutter]{runge_2019}
Runge, F., Stoll, D., Falkner, S., and Hutter, F.
\newblock Learning to design {RNA}.
\newblock In \emph{International Conference on Learning Representations}, 2019.

\bibitem[Taneda(2010)]{taneda2010modena}
Taneda, A.
\newblock Modena: a multi-objective rna inverse folding.
\newblock \emph{Advances and Applications in Bioinformatics and Chemistry},
  pp.\  1--12, 2010.

\bibitem[Tuerk \& Gold(1990)Tuerk and Gold]{tuerk1990systematic}
Tuerk, C. and Gold, L.
\newblock Systematic evolution of ligands by exponential enrichment: Rna
  ligands to bacteriophage t4 dna polymerase.
\newblock \emph{science}, 249\penalty0 (4968):\penalty0 505--510, 1990.

\bibitem[Vorobyeva et~al.(2018)Vorobyeva, Davydova, Vorobjev, Pyshnyi, and
  Venyaminova]{vorobyeva2018key}
Vorobyeva, M.~A., Davydova, A.~S., Vorobjev, P.~E., Pyshnyi, D.~V., and
  Venyaminova, A.~G.
\newblock Key aspects of nucleic acid library design for in vitro selection.
\newblock \emph{International journal of molecular sciences}, 19\penalty0
  (2):\penalty0 470, 2018.

\bibitem[Wachsmuth et~al.(2012)Wachsmuth, Findeiß, Weissheimer, Stadler, and
  Mörl]{wachsmuth_2012}
Wachsmuth, M., Findeiß, S., Weissheimer, N., Stadler, P.~F., and Mörl, M.
\newblock {De novo design of a synthetic riboswitch that regulates
  transcription termination }.
\newblock \emph{Nucleic Acids Research}, 41\penalty0 (4):\penalty0 2541--2551,
  12 2012.
\newblock ISSN 0305-1048.

\bibitem[Zhou et~al.(2015)Zhou, Xia, Luo, Liang, and
  Shakhnovich]{zhou2015searching}
Zhou, Q., Xia, X., Luo, Z., Liang, H., and Shakhnovich, E.
\newblock Searching the sequence space for potent aptamers using selex in
  silico.
\newblock \emph{Journal of chemical theory and computation}, 11\penalty0
  (12):\penalty0 5939--5946, 2015.

\end{thebibliography}
\bibliographystyle{icml2023}

\newpage
\appendix
\onecolumn

\section{Data}
\label{app:data}
\begin{table}[ht]
\centering
\caption[Datasets]{\textbf{Overview of the training and validation sets.}}\vskip 0.15in
\begin{center}
\begin{small}
\begin{sc}
\begin{tabular}{lrrrr}
\toprule
\textbf{Data Set}                   & \textbf{Tasks} &  \textbf{Mean Length} & \textbf{Median Length}      & \textbf{Min/Max Length}   \\

\midrule

Training1 (``long'')            & 100000      & 470.9       & 300            &   200--8033           \\
Training2 (``short'')           & 100000      & 103.9       & 99             &   23--200           \\
Training3 (``random'')           & 100000      & 142.7       & 106            &   23--6361           \\

\midrule

Validation      & 100       & 135.8       & 94            & 46--1802    \\

\bottomrule
\end{tabular}%%
\end{sc}
\end{small}
\end{center}
\vskip -0.1in
\label{tbl:datasets}
\end{table}

Following \citet{runge_2019}, we generate training data from the Rfam database version 14.1 by folding all sequences using \emph{RNAfold}. We build three training sets of 100000 samples each with different length distributions, and a non-overlapping validation set of 100 samples. The datasets are described in Table~\ref{tbl:datasets}.
Instead of specifically training for the task of inverse RNA folding by masking the sequences of the samples only, we first mask up to five parts of the structures, each covering up to 20\% of the total length, while the positions, the lengths, and the number of parts are sampled uniformly at random. In the second step, we mask corresponding parts of the sequences that remain unmasked in the structures to derive tasks of alternating sequences and structure constraints.
Finally, we randomly mask the sequences of $\sim$20\% of the samples to derive tasks that correspond to RNA design from arbitrary sequence and structure parts. Examples of the resulting training tasks are shown in Table~\ref{tbl:task_examples}.

\begin{table}[ht]
\centering
\caption[Task Examples]{\textbf{Examples of different tasks in the training data.}}\vskip 0.15in
\begin{center}
\begin{small}
\begin{sc}
\begin{tabular}{llcr}
\toprule
\textbf{Task Description}                           &\textbf{Task Space}   & \textbf{Example} & \textbf{Percentage of Data} \\ \midrule
\multirow{2}{*}{Inverse RNA Folding}                & Structure            & \texttt{...........((((....))))......} & \multirow{2}{*}{11,5} \\
                                                    & Sequence             & \texttt{?????????????????????????????} & \\
\multirow{2}{*}{Alternating Constraints}            & Structure            & \texttt{?....???.?.((((...???????????} & \multirow{2}{*}{66,7}\\
                                                    & Sequence             & \texttt{C????UAU?C????????UCAGAUAAAAC} & \\
\multirow{2}{*}{Random Masking}                     & Structure            & \texttt{?....???.?.((((...???????????} & \multirow{2}{*}{21,8}\\
                                                    & Sequence             & \texttt{C????U?U?C????????UCAGA????AC} & \\
                                                    
\bottomrule
\end{tabular}%%
\end{sc}
\end{small}
\end{center}
\vskip -0.1in
\label{tbl:task_examples}
\end{table}

\section{Meta-Optimization}\label{app:meta_optimization}

\begin{figure}[htbp]
\vskip 0.2in
\begin{center}
        \centerline{%
    \includegraphics[width=\linewidth]{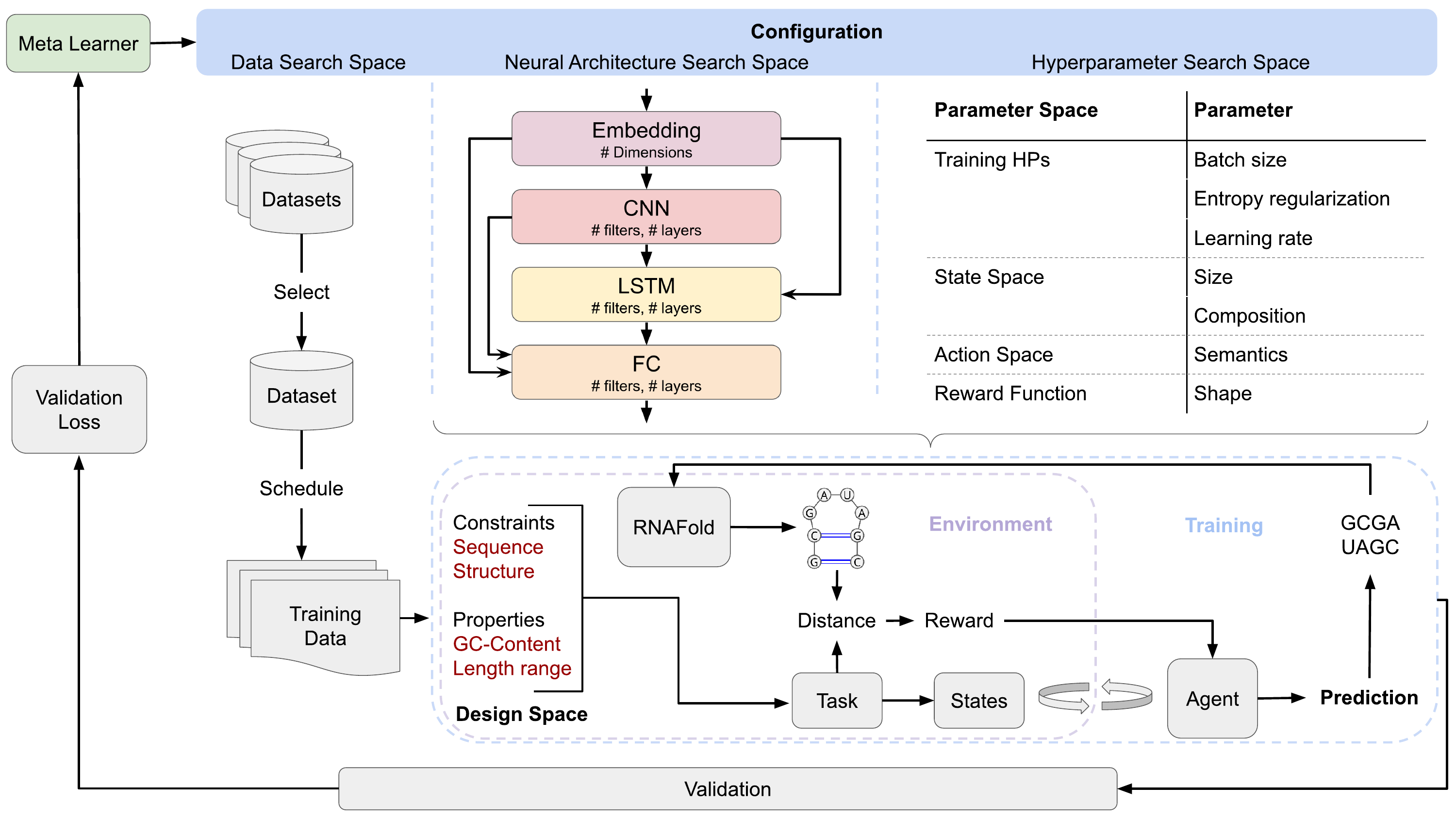}
    }
    \caption[Meta-Optimization Loop]{\textbf{Meta-optimization loop.} In each iteration, the meta-learner samples a configuration from a rich configuration space. The sampled configuration defines the learning algorithm's hyperparameters, a specific environment, a particular network architecture of the agent, a training curriculum, and a training data set. These components together formulate the learner; a deep reinforcement learning algorithm, which is trained on the selected training set and evaluated on a validation set. The resulting validation loss is communicated to the meta-learner to update its model, seeking to learn to sample better configurations with each iteration.}
    \label{fig:meta_optimization_loop}
\end{center}
\vskip -0.2in
\end{figure}

We use an automated deep reinforcement learning (AutoRL) approach that automatically selects the best RL setting for \emph{libLEARNA}, given a rich configuration space as proposed by \citet{runge_2019}. In particular, we use a meta-optimization process to jointly optimize the formulation of our RL algorithm: in the outer loop, the meta-learner iteratively samples a configuration that defines an RL algorithm, which is then used to learn an RNA design policy in the inner loop. The resulting policy is evaluated on a validation data set and the meta-learner observes the validation loss to update its own model accordingly.
The meta-optimization loop for \emph{libLEARNA} is illustrated in Figure~\ref{fig:meta_optimization_loop}.

For \emph{libLEARNA}, we mainly adopt the configuration space proposed by \citet{runge_2019} but introduce four novel dimensions, described in the following.

\paragraph{Action Semantics} Inverse RNA folding algorithms typically benefit from directly predicting Watson-Crick base pairs at paired sites. For \emph{libLEARNA}, however, many paired sites are not trivially distinguishable and we included a dimension via the \emph{action semantics} parameter to either use direct prediction of pairs or not, to potentially exploit predictions of non-Watson-Crick base pairs.

\paragraph{Training Data} The choice of training data and the schedule of the tasks during training can have a strong impact on the final performance of the learning algorithm. We, therefore, include two dimensions to choose from different training data distributions (\emph{training data} parameter) and curricula (\emph{curriculum} parameter).

\paragraph{Individual State Composition} The predictions of the RL agent are based on the presented states. While the formulation of individual states in \citet{runge_2019} was based only on the provided target structure, we decide to add a choice to formulate states that provide the agent with information about its own decisions. We use the \emph{state composition} parameter $\sigma$ to choose either to define states based on the target task or to define states using the design process of the agent (updating the representation at each time step).

Overall, our design choices yield an 18-dimensional configuration space, encompassing a broad range of neural architectures to formulate the agent (including elements of recurrent neural networks (RNNs) and convolutional neural networks (CNNs)), a variety of different environment formulations, three distinct training data distributions, two training curricula, and training hyperparameters.
The complete list of parameters, their types, ranges, and the priors we used over them as well as the final selected configuration of \emph{libLEARNA} are listed in Table~\ref{tab:configuration_space}.

\begin{table}[!ht]
\centering
\caption[Final Configurations]{\textbf{Configuration space and finally selected parameters of \emph{libLEARNA}.}
 }
\begin{center}
\begin{small}
\begin{sc}
\centering
\begin{tabular}{llcrr}
\toprule
\textbf{Parameter Name} & \textbf{Type} & \textbf{Range} & \textbf{Prior} & \textbf{libLEARNA} \\ \midrule
state radius $\kappa$&integer&[$0$,\, $32$]&uniform&10\\
individual state composition $\sigma$ &categorical&[``target'', ``design'']&uniform&``target''\\
action semantics&categorical&[``pair'', ``single'']&uniform&``pair''\\
reward exponent $\alpha$&float&[$1$,\, $12$]&uniform&10.76 \\
filter size in $1^{\text{st}}$ conv layer&integer& $\{ 0 \} \cup \{3,\, 5,\, \ldots,\, 17 \}$ &uniform&0\\
filter size in $2^{\text{nd}}$ conv layer&integer& $\{ 0,\, 3,\, 5,\, 7,\, 9 \}$ &uniform&0\\
\# filter in $1^{\text{st}}$ conv layer&integer&[$1$,\, $32$]&log-uniform&17\\
\# filter in $2^{\text{nd}}$ conv layer&integer&[$1$,\, $32$]&log-uniform&24\\
\# LSTM layers&integer&[$0$,\, $3$]&uniform&0\\
\# units in every LSTM layer&integer&[$1$,\, $64$]&log-uniform&20\\
\# fully connected layers&integer&[$1$,\, $2$]&uniform&2\\
\# units in fully connected layers &integer&[$8$,\, $64$]&log-uniform&12\\
embedding dimensionality&integer&[$0$,\, $21$]&uniform&17\\
batch size&integer&[$32$,\, $256$]&log-uniform&247\\
entropy regularization&float&[$1 \cdot 10^{-7}$,\, $1\cdot 10^{-2}$]&log-uniform&$4.46\cdot 10^{-7}$\\
learning rate for PPO&float&[$1 \cdot 10^{-6}$,\, $1\cdot 10^{-3}$]&log-uniform&$5.9\cdot 10^{-4}$ \\
training data&categorical&[``random'', ``short'', ``long'']&uniform&``short''\\
training curriculum&categorical&[``random'', ``sorted'']&uniform&``sorted''\\
\bottomrule
\end{tabular}%
\end{sc}
\end{small}
\end{center}
\vskip -0.1in
\label{tab:configuration_space}
\end{table}

\section{Improvement Steps}
\label{app:GIS}
\emph{LEARNA} employs a local improvement step (LIS) to exhaustively search over neighboring sequences if the agent is close to a solution for a given task to account for the stochasticity of the agent. We adopt this procedure to improve the structure loss for a given task. During evaluations, we add a desired GC-content to the task. We, therefore, develop a GC-improvement step (GIS) to guide the agent, either increasing or decreasing the GC content of the designed sequence by replacing nucleotides at random positions with their respective counterparts (either A or U by G or C if the GC-content is too low, or vice-versa). The GIS becomes active whenever the LIS becomes active and changes to the sequence are only applied if the structure loss is at least as low as before the GIS.

\end{document}